\documentclass[11pt]{article}
\pdfoutput=1
\usepackage[margin=1in]{geometry}
\usepackage{cprotect}
\usepackage{color}
\usepackage[colorlinks=true,linkcolor=blue,citecolor=black,urlcolor=black]{hyperref}
\usepackage{amssymb,amsmath}
\usepackage{amsthm}

\usepackage{graphicx}
\usepackage{helvet}
\usepackage[font=footnotesize,labelfont=bf]{caption}
\usepackage{longtable}
\usepackage{url}
\usepackage{subcaption}

\usepackage{multirow}
\usepackage{ulem}
\usepackage{makecell}
\usepackage{tcolorbox}

\usepackage{titlesec}

\titleformat*{\section}{\LARGE\bfseries}
\titleformat*{\subsection}{\Large\bfseries}
\titleformat*{\subsubsection}{\large\bfseries}

\usepackage[square,sort&compress,numbers]{natbib}
\usepackage[T1]{fontenc}
\usepackage[utf8]{inputenc}
\usepackage{tabularx,ragged2e,booktabs,caption}
\newcolumntype{C}[1]{>{\Centering}m{#1}}

\newcommand\blfootnote[1]{%
  \begingroup
  \renewcommand\thefootnote{}\footnote{#1}%
  \addtocounter{footnote}{-1}%
  \endgroup
}

\DeclareMathOperator*{\argmin}{arg\,min}
\usepackage{setspace}
\numberwithin{equation}{section}
\theoremstyle{plain}

\usepackage{mathtools}

\usepackage{array}

\usepackage{amssymb}
\usepackage{pifont}
\newcommand{\R}{{\rm I}\kern-0.18em{\rm R}}

\title{Particle Swarm Optimization for Constrained Maximum Likelihood Estimation: A Case Study in Pseudotime Analysis}

\author{%
Elvis Cui\,$^{1}$, Dongyuan Song\,$^{2}$, Weng Kee Wong\,$^{1,*}$
}

\date{\vspace{-5ex}}

\begin{document}
\maketitle
\blfootnote{\\$^{1}$ Department of Biostatistics, University of California, Los Angeles, CA\\
$^{2}$ Department of Statistics, University of California, Los Angeles, CA\\
$^{*}$ To whom correspondence should be addressed.
Email: wkwong@ucla.edu
}

\begin{abstract}
    The aim of paper is to apply two types of particle swarm optimization, global best and local best PSO to a constrained maximum likelihood estimation problem in pseudotime analysis, a sub-field in bioinformatics. The results have shown that particle swarm optimization is extremely useful and efficient when the optimization problem is non-differentiable and non-convex so that analytical solution can not be derived and gradient-based methods can not be applied.
\end{abstract}


\section{Introduction}
Parametric statistical models are commonly used in many sub-fields of bioinformatics \cite{li2018modeling}, \cite{li2019statistical}. For simplicity and computational concerns, bioinformatic scientists prefer to use differentiable and unconstrained statistical models than non-differentiable and constrained ones. For example, in pseudotime analysis (see section 3), in  \cite{trapnell2014dynamics}, the authors propose to regress gene expression on pseudotime using cubic B-spline so that an analytical solution is available. Other authors suggest to replace B-spline with a generalized linear model and a gradient-based method is applied to find maximum likelihood estimation \cite{campbell2017switchde}. In zero imputation problem, the authors construct a Gamma-Normal mixture model so that parameters can be estimated analytically \cite{li2018accurate}. In \cite{jiang2020mbimpute}, the authors propose an unconstrained LASSO-type objective function and optimize it with a convex optimization algorithm. 

However, in real applications, it is common to impose constraints on parameters for interpretability. Besides, analytically solutions are not always available and the likelihood function is not differentiable or convex if discrete parameters are contained. Thus, constrained models without desirable mathematical properties can be more realistic and interpretable in many cases.  For instance, the projection matrix is imposed to be non-negative in \cite{song2021scpnmf} so that genes with positive values will be selected. Yet, optimization with respect to these models is a stumbling block for scientists because many optimization algorithms are only applicable when gradients are available and when the problem is unconstrained.

Motivated by problems described above, in this paper we illustrate how to apply particle swarm optimization, a optimization algorithm introduced in next section, to solve a constrained maximum likelihood estimation problem in pseudotime analysis in section 3. The objective function of the problem is non-differentiable, non-convex and has domain constraints so that it cannot be minimized directly using gradient-based methods such as L-BFGS-B, Newton-Raphson. On the contrary, as shown in section 4, PSO can produce nearly optimal solutions within only a few iterations.

The rest of the paper is organized as follows. Section 2 introduces two PSO algorithms, global best and local best PSO. Section 3 introduces a probabilistic model with boundary constraints in pseudotime analysis. In section 4, we apply both PSOs to perform a simulation study with different settings. In section 5, we draw conclusions of the paper.

\section{Particle Swarm Optimization}

Swarm intelligence algorithms, such as ant colony algorithms \cite{dorigo2006ant}, cuckoo search algorithms \cite{yang2009cuckoo} and firefly algorithms \cite{yang2009firefly}, are mimicking the behaviour of a swarm to solve optimization problems. They are now receiving more and more interest and attention not only in the literature of mathematics, but also in econometrics, optimal design, engineering, etc \cite{yang2017nature}. Particle swarm optimization (PSO), which was proposed by Kennedy and Eberhart in 1995 \cite{kennedy1995particle}, is one of the most widely used swarm intelligence algorithms to optimize an objective function with boundary constraints. It is the main optimization tool in this paper and is introduced in below.

PSO solves the problem by producing a sequence of candidate solutions. Unlike gradient descent algorithms in deep learning, PSO does not require either differentiability or the convexity \cite{boyd2004convex} of the objective function and constraints. Therefore, PSO is particularly useful when the objective function does not have desirable analytical properties or the derivative of the objective function does not exist. Subsection 2.1 introduces the PSO algorithm with global best topology and how it works. Subsection 2.2 introduces the local best topology as a variation of global best PSO algorithm.

\subsection{PSO Algorithm}

PSO encodes swarm intelligence, such as bird flocking, fish schooling, into two simple dynamic equations to solve optimization problems of the following form:
\begin{align*}
    \min\ &f(x)\\
    \text{s.t. } &x\in\mathcal{S}
\end{align*}

Where $x\in\mathbb{R}^d$ is a $d$-dimensional vector, $f(x)$ is a real-valued objective function (measurability is the only requirement) and $\mathcal{S}\subset\mathbb{R}^d$ is the search space or domain of $x$. The algorithm starts with $n$ alternative values of $x$, denoted as $x_1^{0},\cdots, x_n^{0}$. Each $x_i^{0},i=1,\cdots n$ represents a \emph{particle} and is initialized with a velocity $v_i^{0}\in\mathbb{R}^d$. Then for $i=1,\cdots,n$, PSO iterates with the following two equations \cite{bratton2007defining}:
\begin{align}\label{eq:pso}
    v_{i}^{k+1} &= wv_i^{k}+c_1r_{i1}^k(p_i^k-x_i^k)+c_2r_{i2}^k(p_g^k-x_i^k)\nonumber\\
    x_{i}^{k+1}&=x_i^k+v_i^{k+1}
\end{align}
Where $k=0,1,\cdots$ is the number of iteration, $w$ is called the \emph{inertia weight}, $c_1$ and $c_2$ are called \emph{cognitive} and \emph{social} parameters respectively and $r_{i1}^k$, $r_{i2}^k$ are two independent random numbers distributed uniformly in $[0,1]$. Usually $w, c_1$ and $c_2$ are set to numbers between $[0,2]$ by users. Most importantly,
\begin{align*}
    p_{i}^k&=\argmin_{x\in A_i} f(x)\\
    p_g^k&=\argmin_{x\in\cup_{i=1}^n A_i} f(x) \\
    A_i&=\{x_i^t:t=0,\cdots,k\}
\end{align*}
Thus, $p_i^k$ is the best position recorded by particle $i$ up to $k^{th}$ iteration and $p_g^k$ is the best position recorded by the whole swarm up to $k^{th}$ iteration. The inertia weight $w$ controls the level of a particle moving towards its last direction $v_i^k$. The cognitive parameter $c_1$ represents how a particle is affected by its best known position $p_i^k$. Similarly, the social parameter $c_2$ determines the influence of the swarm's best knowledge $p_g^k$ on particle $i$.

Because $p_g^k$ is the best solution found by the whole swarm, the set of equations \ref{eq:pso} is also called \emph{global best PSO} (gbest PSO)  \cite{bratton2007defining}. In next subsection, we introduce a variant of gbest PSO known as \emph{local best PSO} (lbest PSO), which replaces $p_g^k$ by a locally optimal solution.

\begin{figure}[htp]
  \centering
  \caption{Illustration of PSO.}
  \begin{tabular}{c}

   \includegraphics[width=18cm]{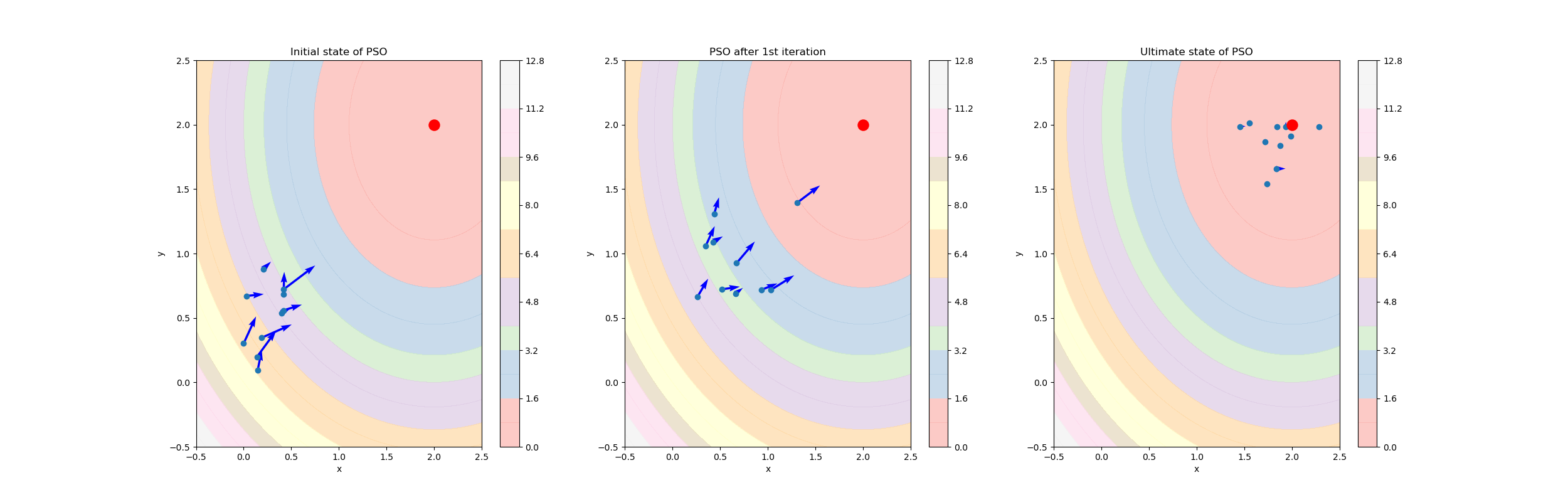}

  \end{tabular}
  \label{fig:illu}

\end{figure}

To better understand the logic of PSO, suppose we have 10 ants starting around origin $(0,0)$ and they are looking for food at point $(2,2)$ (left panel of figure \ref{fig:illu}). Colors in background represent distances to the food. The initial position of each ant corresponds to $x_i^0$ and the objective function $f(x)$ is the Euclidean distance between point $x$ and the food at $(2,2)$. Each ant is initialized with an velocity $v_i^0$ (blue arrow). After moving one step, ants re-analyse their positions and the distances to the food so that (1) the best position of an individual is recorded and this is $p_i^1$; (2) the best position of all ants is recorded, which is $p_g^1$. Here \emph{best} refers to the minimum distance to the food at (2,2). Thus, each of them re-corrects its velocity according to equation \ref{eq:pso} (middle panel of figure \ref{fig:illu}). After several iterations, all ants gather around the food and the velocity decreases to 0 gradually (right panel of figure \ref{fig:illu}).

\bigskip

\subsection{Topology of PSO}
Recall that in \ref{eq:pso}, we have $p_g^k=\argmin_{x\in\cup_{i=1}^nA_i}f(x)$ and $p_g^k$ is called global best of the swarm. For this reason, the topology of gbest PSO is called \emph{gbest topology} in literature. However, having global best knowledge sometimes converges too fast and leads to sub-optimal point. Thus, \emph{local best PSO} (lbest PSO) \cite{eberhart1995new} is proposed as a remedy for the sub-optimal problem of gbest PSO and the structure of particles is referred as \emph{local topology} in literature. 

Instead of finding the global minimum among all particles, lbest model replace $p_g^k$ in \ref{eq:pso} with
\begin{align*}
    p_{l}^k&=\argmin_{x\in \cup_{i\in B_i^k}A_i}f(x)\\
    A_i&=\{x_i^t:t=0,\cdots,k\}\\
    B_i^k&=\{j:\text{ particle $j$ is a neighbor of $i$.}\}
\end{align*}
In other words, particle $i$ only has information among its neighbors. A popular choice of neighbor is the Euclidean topology. That is, in $k^{th}$ iteration, particle $j$ is a neighbor of $i$ if the rank of the Euclidean distance from $j$ to $i$ is less than a pre-specified number, say, 5. In this paper, we use gbest model and lbest model with Euclidean topology to solve optimization problems.

\section{A Probabilistic Model with Boundary Constraints}
The task of single-cell pseudotime analysis is to map gene expression profiles to a unique value called \emph{pseudotime} which represents the progression of a biological process \cite{trapnell2014dynamics}. After a pseudotime has been assigned to each cell we may identify genes that has a strong correlation with pseudotime through differential expression testing \cite{campbell2017switchde}. In the following, we model the correlation between pseudotime and gene expression via a parametric statistical model.

Let $Y$ be the $C\times G$ expression matrix for $G$ genes and $C$ cells with column vector $y_g,g=1,\cdots,G$. Let $\{t_c:c=1,\cdots,C\}$ be the set of the pseudotime of cell $c=1,\cdots,C$, and we write $y_g^T=(y_{g1},y_{g2}, \cdots,y_{gC})$. Using the mean parametrization of negative binomial distribution, we assume at each pseudotime $t_c$, the marginal distribution of $y_g$ is
\begin{align}\label{eq:link_fct}
    y_{gc}&\sim \text{NB}(\tau_{t_c},\phi), g=1,\cdots,G\nonumber\\
    \tau_{t_c}&=\frac{2\mu_g}{1+\exp(-k_g(t_c-t_g))}
\end{align}
where $\phi_g, \mu_g, k_g, t_g$ are parameters to be estimated. They can be interpreted as:
\begin{itemize}
    \item $\mu_g:$ average peak expression.
    \item $k_g:$ activation strength or how quickly a gene is up-or-down regulated.
    \item $t_g:$ activation time or where in the trajectory the gene regulation occurs.
    \item $\phi_g: $ parameter of negative binomial distribution.
\end{itemize}
The positive and negative sign of $k_g$ indicate the monotone property of $\tau_{t_c}$ in $t_c$. The value of $t_g$ determines the turning point of $\tau_{t_c}$, a larger $t_g$ corresponds to a turning point closer to $1$. $\mu_g$ controls the height of $\tau_{t_c}$ and $\phi$ controls slightly the shape of $\tau_{t_c}$. Hence the likelihood function for $\theta^T=(k_g,t_g,\mu_g,\phi_g)$ is
\begin{align}\label{eq:likelihood}
    L(\theta|y_g)&=\prod_{c=1}^C\mathbb{P}_{Y_{gc}}(Y_{gc}=y_{gc})\nonumber\\
    &=\prod_{c=1}^C\binom{y_{gc}+\phi_g-1}{y_{gc}}\left(\frac{\tau_{t_c}}{\tau_{t_c}+\phi_g}\right)^{y_{gc}}\left(\frac{\phi_g}{\tau_{t_c}+\phi_g}\right)^{\phi_g}
\end{align}
where $\tau_{t_c}$ is given by equation \ref{eq:link_fct}. Besides, to improve intepretability, we also assume that $2\mu_{g}$ does not exceed the maximum of $y_{gc},c=1,\cdots,C$ and $t_g$ should be within the range of $t_c,c=1,\cdots,C$. To sum up, estimating the parameters $\theta_g$ for $g=1,\cdots,G$ boils down to the following constrained optimization problem:
\begin{align}\label{eq:cons_optim}
    \min_{\theta} &-\log L(\theta|y_g)\nonumber\\
    \text{s.t. }&\min_{c}y_{gc}\le2\mu_g\leq \max_{c}y_{gc}\nonumber\\
    &\min_ct_c\le t_g\le\max_ct_c
\end{align}

In practice, $t_c$'s are always scaled within range $[0,1]$ so $t_g$ has constraint $0< t_g< 1$. The function \ref{eq:likelihood} is not differentiable in $\phi$ and the gradient behaves erratically in $k_g$, $\mu_g$ and $t_g$. Thus, gradient-based optimization methods are not applicable in this case.

\section{Simulation study}
In this section, we apply two PSOs to tackle the optimization problem \ref{eq:cons_optim} using simulated data.
\subsection{Simulation setting}
Without loss of generality, we generate the data using model \ref{eq:link_fct} with $G=1$ and  6 different settings of parameters in table \ref{tab:simu}. For instance, setting 1 means $Y=(y_{1,1},\cdots,y_{1,400})^T$ is a $400\times 1$ expression vector and each $y_{1c}$ is independently generated from a negative binomial model with parameters $\tau_{t_c}$ and $\phi_g=25$ and
$$\tau_{t_c}=\frac{2\times 6}{1+\exp(7(t_c-0.4))}$$

Given $C$, each $t_c$ is generated independently and randomly within interval $[0,1]$. The different choices of parameters in \ref{tab:simu} is motivated by empirical evidence of scRNA-seq data.

\bigskip

\begin{minipage}{\linewidth}

\centering

\captionof{table}{Simulation setting-up of the regression model.} \label{tab:simu} 
\begin{tabular}{ C{.75in} C{.55in} *4{C{.75in}}}\toprule[1.5pt]
\bf Setting & \bf $C$  & \bf $k_g$ & \bf $t_g$ & \bf $\mu_g$ & \bf $\phi_g$\\\midrule
\bf 1 & 400  & 7 & 0.4 & 6 & 25 \\\midrule
\bf 2 & 400  & -8 & 0.85&4 & 80 \\\midrule
\bf 3 & 400  & 1.6 & 1&1.4 & 2 \\\midrule
\bf 4 & 100  & 7 & 0.4&6 & 25 \\\midrule
\bf 5 & 100  & -8 & 0.85&4 &80 \\\midrule
\bf 6 & 100  & 1.6& 1 &1.4& 2 \\
\bottomrule[1.25pt]
\end {tabular}\par
\bigskip
\end{minipage}

There are 6 different datasets and each dataset contains $C$ pairs of values:
$$\{(t_{c},y_{gc}):c=1,2,\cdots,C\}$$

where $g=1$ and $y_{gc}$'s are generated from model \ref{eq:link_fct}. There are $(400\times3+100\times 3)=1,500$ pairs of values in total.

\subsection{Tuning parameter setting} There are three tuning parameters in gbest PSO and they are $w$, the inertia weight, $c_1$, the cognitive parameter and $c_2$, the social parameter. By prior knowledge, we know that $t_g\in[0,1]$, $k_g$ can be either positive and negative, $\mu_g$ depends on the range of $y_c$ and $\phi$ can vary from small to large. Thus, $c_1$ should be large compare to $c_2$ and $w$ should be moderate, suggesting that particles intend to search in a relatively larger space instead of concentrating on a small area. Thus, for gbest, we set

$$(w,c_1,c_2)=(0.9,1.5,0.3)$$

for all simulation settings. For lbest PSO, we set $m$, the number of neighbors, to be $5$. Thus, for lbest, we set

$$(m,w,c_1,c_2)=(5,0.9,1.5,0.3)$$

Both PSOs are run $50$ times with $100$ iterations and $10$ particles to ensure a reasonable statistical result to compare gbest and lbest.

\subsection{Simulation results}
Table \ref{tab:result_pso} shows the results of gbest PSO and lbest PSO solving \ref{eq:cons_optim}, respectively. The bold numbers refer the better solutions, i.e. the lower negative log likelihood function values found by PSO, in each setting. Although gbest PSO outperforms lbest PSO in almost all settings, lbest PSO has lower standard deviation (std) when the number of pairs is lower (setting 4,5,6 and $C=100$). Since mean, median and std are three measurements of the robustness of optimization algorithms, the results suggest that lbest PSO is more robust when $C=100$ while gbest PSO is more reliable in the case $C=400$.
\bigskip

\begin{minipage}{\linewidth}

\centering

\captionof{table}{Results of gbest and lbest PSO in different settings based on \ref{eq:cons_optim}.} \label{tab:result_pso} 
\begin{tabular}{ C{.65in} *8{C{.5in}}}
\toprule[1.5pt]
&\multicolumn{4}{|c|}{gbest}&\multicolumn{4}{|c}{lbest}\\
\toprule[1.5pt]
\bf Setting & \bf best & \bf mean & \bf std & \bf median & \bf best & \bf mean & \bf std & \bf median\\\midrule
\bf 1 & \bf965.85 & 969.32 & 5.20 & 966.44 & 966.11 & 974.01 & 9.55 & 970.11 \\\midrule
\bf 2 & \bf919.19 & 941.42 & 16.24 & 941.10 & 920.17 & 950.23 & 12.97 & 951.84 \\\midrule
\bf 3 & \bf539.33 & 540.21 & 1.60 & 539.58 & 539.36& 540.75 & 2.07 & 539.83  \\\midrule
\bf 4 & \bf246.77 & 248.32 & 3.33 & 247.10 & \bf246.77 & 248.80 & 2.31 & 248.19 \\\midrule
\bf 5 & \bf233.24  & 234.90 & 2.57 & 233.89 &233.26 & 234.49 & 2.12 & 233.93 \\\midrule
\bf 6 & \bf141.35 & 141.74& 0.37& 141.57 & 141.39 & 141.83 & 0.36 & 141.61 \\
\bottomrule[1.25pt]
\end {tabular}\par
\bigskip
\end{minipage}

Table \ref{tab:para_est} shows the best parameter estimation results of problem \ref{eq:cons_optim}. The parameter estimation results returned by gbest PSO and lbest PSO are similar to each other. In addition, except for setting 6, both PSOs give consistent estimation of parameters compared with the true value in table \ref{tab:simu}. It suggests that the inconsistent estimation in setting 6 is due to maximum likelihood estimation itself instead of optimization algorithms.

\bigskip

\begin{minipage}{\linewidth}

\centering

\captionof{table}{The best parameter estimation results based on \ref{eq:cons_optim}.} \label{tab:para_est} 
\begin{tabular}{ C{.65in} *8{C{.52in}}}
\toprule[1.5pt]
&\multicolumn{4}{|c|}{gbest}&\multicolumn{4}{|c}{lbest}\\
\toprule[1.5pt]
\bf Setting & \bf $k_g$ & \bf $t_g$ & \bf $\mu_g$ & \bf $\phi_g$& \bf $k_g$ & \bf $t_g$ & \bf $\mu_g$ & \bf $\phi_g$\\\midrule
\bf 1 & 5.4294 & 0.4655 & 6.3844 & 17 & 5.1776 & 0.4833 & 6.5585 & 16 \\\midrule
\bf 2 & -9.7579 & 0.8610& 3.9048& 66 & -8.3863 & 0.8398 & 3.9933 & 60 \\\midrule
\bf 3 & 1.6199 & 0.9789 & 1.5215 & 2 & 1.6983 & 0.9364 & 1.4810 & 2 \\\midrule
\bf 4 &  6.4729 & 0.3814 & 6.0773 & 15& 6.5545 & 0.3781 & 6.0621&15  \\\midrule
\bf 5 & -3.7905 & 0.7915 & 4.6435 & 76 & -3.6032 & 0.7818 & 4.7309 & 77 \\\midrule
\bf 6 & 2.7108  & 0.2148 & 0.8032 & 2 &3.0299 & 0.2688 & 0.8193 & 3 \\
\bottomrule[1.25pt]
\end {tabular}\par
\bigskip
\end{minipage}

\bigskip

\begin{figure}
\begin{subfigure}{.5\textwidth}
  \centering
  \includegraphics[width=1.\linewidth]{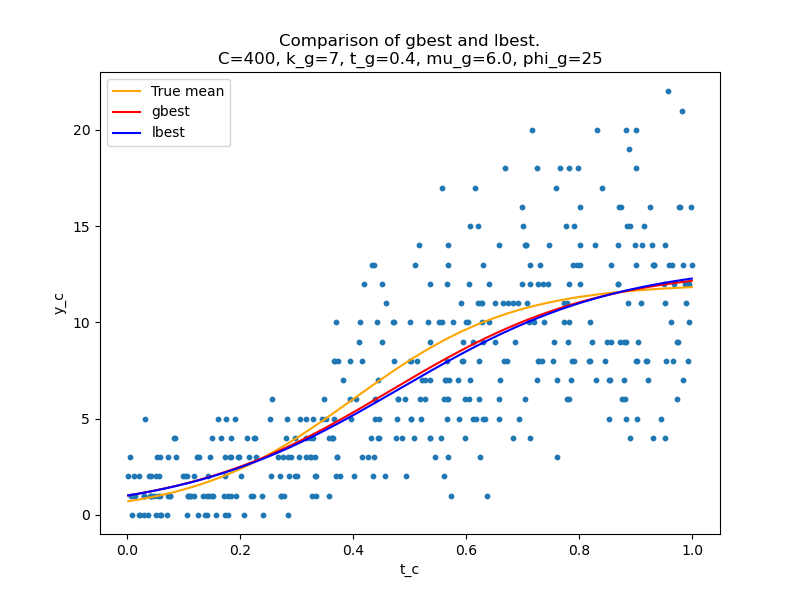}
  \caption{Setting 1}
  \label{fig:sfig1}
\end{subfigure}%
\begin{subfigure}{.5\textwidth}
  \centering
  \includegraphics[width=1.\linewidth]{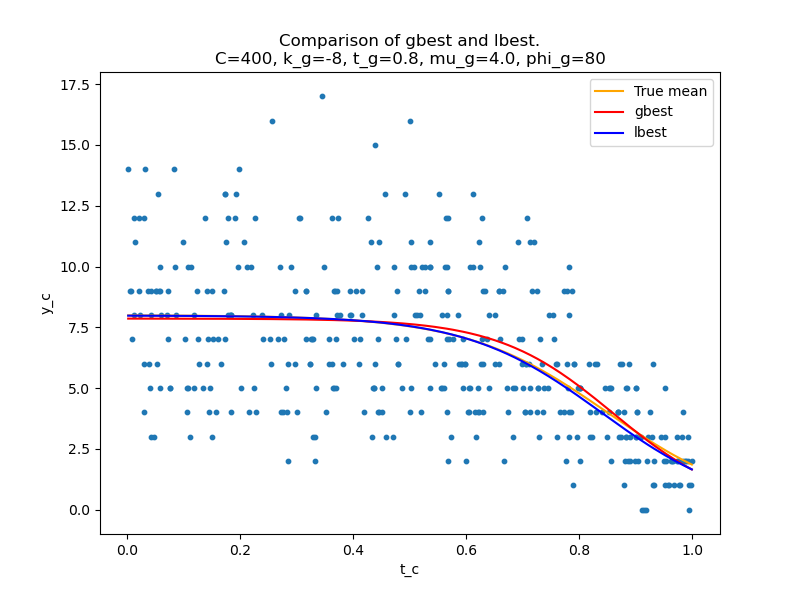}
  \caption{Setting 2}
  \label{fig:sfig2}
\end{subfigure}\\
\begin{subfigure}{.5\textwidth}
  \centering
  \includegraphics[width=1.\linewidth]{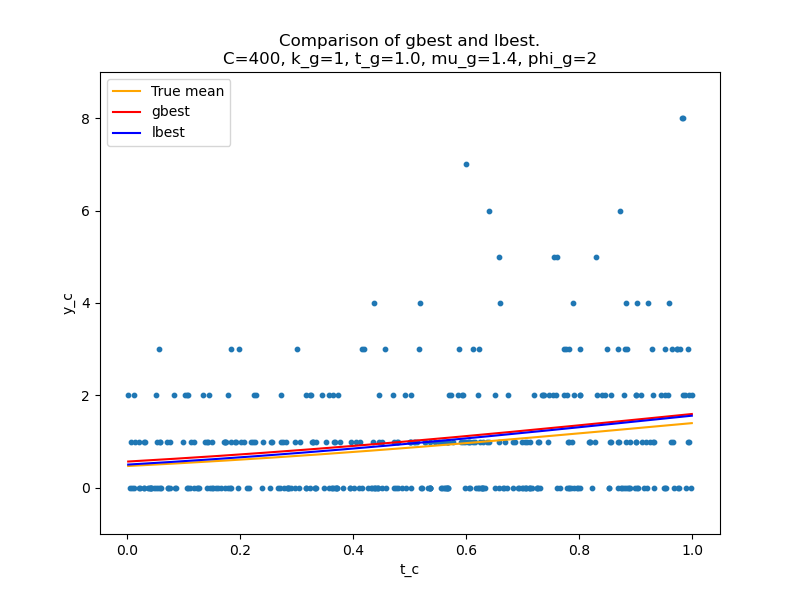}
  \caption{Setting 3}
  \label{fig:sfig1}
\end{subfigure}%
\begin{subfigure}{.5\textwidth}
  \centering
  \includegraphics[width=1.\linewidth]{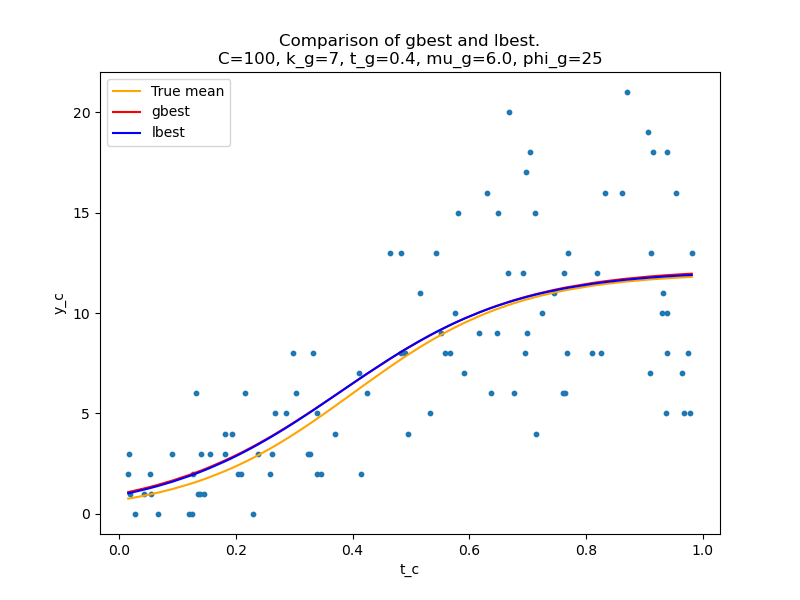}
  \caption{Setting 4}
  \label{fig:sfig2}
\end{subfigure}\\
\begin{subfigure}{.5\textwidth}
  \centering
  \includegraphics[width=1.\linewidth]{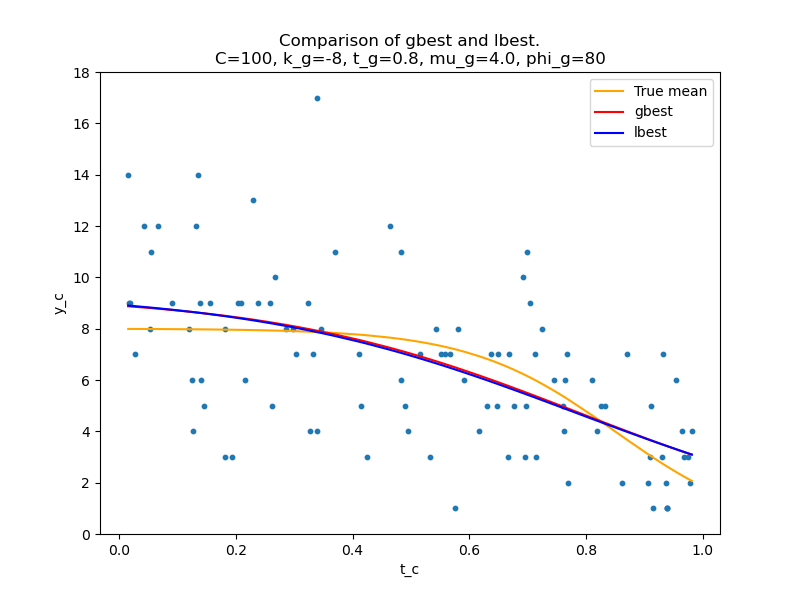}
  \caption{Setting 5}
  \label{fig:sfig1}
\end{subfigure}%
\begin{subfigure}{.5\textwidth}
  \centering
  \includegraphics[width=1.\linewidth]{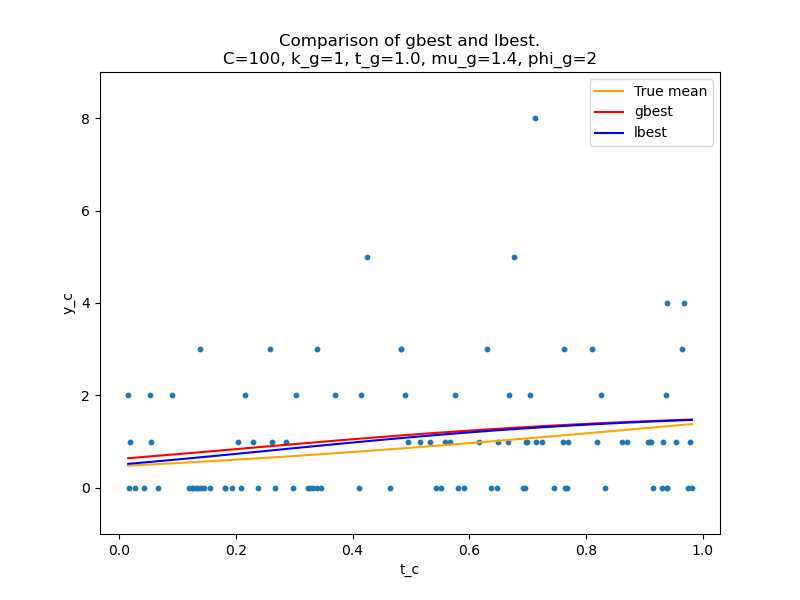}
  \caption{Setting 6}
  \label{fig:sfig2}
\end{subfigure}
\caption{The comparison of regression results and simulated data.}
\label{fig:comparison}
\end{figure}

Each subplot of figure \ref{fig:comparison} shows the fitted $\tau_{t_c}$ using estimated parameters as well as simulated data. The true $\tau_{t_c}$ curve is in orange, the red and blue curve correspond to fitted $\tau_{t_c}$ by gbest and lbest, respectively. Blue points are simulated $y_c$ using model \ref{eq:link_fct}.

From all 6 plots, we see that both gbest and lbest give reasonable results compare with the true value, suggesting that they are efficient in finding optimal solutions of such pseudotime analysis problems.

\section{Conclusion}
The conventional gradient-based methods such as BFGS, gradient descent are not suitable to solve real world optimization problems, because the objective function can be non-differentiable (e.g. $\phi_g$) and the search space is bounded (e.g. $t_g$ and $\mu_g$). In contrast, the swarm intelligence algorithm is extremely useful and efficient solving the constrained optimization problems presented above. 

In this paper, we apply two types of PSO, gbest and lbest, to solve a constrained maximum likelihood problem in pseudotime analysis. With only 10 particles and a few iterations, both gbest and lbest PSO algorithms are able to find the nearly optimal solution of a real world optimization problem. The experimental results also indicate that, though differnces in minimum, std and mean, both algorithms give reasonable estimations and the fitted curve is consistent with the true one.

We hope that this paper can benefit students and scholars in statistics, bioinformatics and other related fields facing different types of optimization problems.

\newpage
\section*{Codes and Datasets}\label{codes}

For codes and datasets used in this paper, please see

\url{https://github.com/ElvisCuiHan/ParticleSwarmOptimization}.

\section*{Acknowledge}

\bibliography{references}

\end{document}